\documentclass[runningheads]{llncs}
\usepackage{lineno,hyperref}
\modulolinenumbers[5]
\usepackage{graphicx}
\usepackage{tcolorbox}
\usepackage{ulem}
\usepackage{tabularx}
\usepackage{soul}
\usepackage[T1]{fontenc}
\usepackage{commath}
\usepackage{booktabs}
\usepackage{todonotes} 
\usepackage{multirow}
\begin{document}
\title{Using a KG-Copy Network\\ for Non-Goal Oriented Dialogues}
%
%
\author{Debanjan Chaudhuri\inst{1,2} \and
Md Rashad Al Hasan Rony\inst{1,2} \and 
Simon Jordan\inst{3} \and
Jens Lehmann\inst{1,2}}
\authorrunning{Chaudhuri et al.}
%
\institute{Enterprise Information Systems Department, Fraunhofer IAIS, Dresden and St.~Augustin, Germany \\
\email{\{debanjan.chaudhuri, md.rashad.al.hasan.rony, jens.lehmann\}}@iais.fraunhofer.de \\
\url{http://iais.fraunhofer.de}
\and
Smart Data Analytics Group, University of Bonn, Germany
\email{\{chaudhur, jens.lehmann\}}@cs.uni-bonn.de, \email{s6mdrony@uni-bonn.de}\\
\url{http://sda.tech} 
\and
Volkswagen Group Research, Wolfsburg, Germany\\
\email{simon.jordan@volkswagen.de}
}
\maketitle              
\begin{abstract}

Non-goal oriented, generative dialogue systems lack the ability to generate answers with grounded facts. A knowledge graph can be considered an abstraction of the real world consisting of well-grounded facts.
This paper addresses the problem of generating well-grounded responses by integrating knowledge graphs into the dialogue system's response generation process, in an end-to-end manner. A dataset for non-goal oriented dialogues is proposed in this paper in the domain of soccer, conversing on different clubs and national teams along with a knowledge graph for each of these teams. A novel neural network architecture is also proposed as a baseline on this dataset, which can integrate knowledge graphs into the response generation process, producing well articulated, knowledge grounded responses. Empirical evidence suggests that the proposed model performs better than other state-of-the-art models for knowledge graph integrated dialogue systems. 

\keywords{Non-goal oriented dialogues  \and knowledge grounded dialogues \and knowledge graphs.}
\end{abstract}
\section{Introduction}

With the recent advancements in neural network based techniques for language understanding and generation, there is an upheaved 
interest in having systems which are able to have articulate conversations with humans. Dialogue systems can generally be classified into goal and non-goal oriented systems, based on the nature of the conversation. The former category includes systems which are able to solve specific set of tasks for users within a particular domain, e.g. restaurant or flight booking. Non-goal oriented dialogue systems, on the other hand, are a first step towards chit-chat scenarios where humans engage in conversations with bots over non-trivial topics. Both types of dialogue systems can benefit from added additional world knowledge \cite{chaudhuri2018improving},\cite{eric2017keyval},\cite{zhu2017flexible}. 

\begin{figure}
    \centering
    \vspace{-1cm}
    \includegraphics[width=\textwidth]{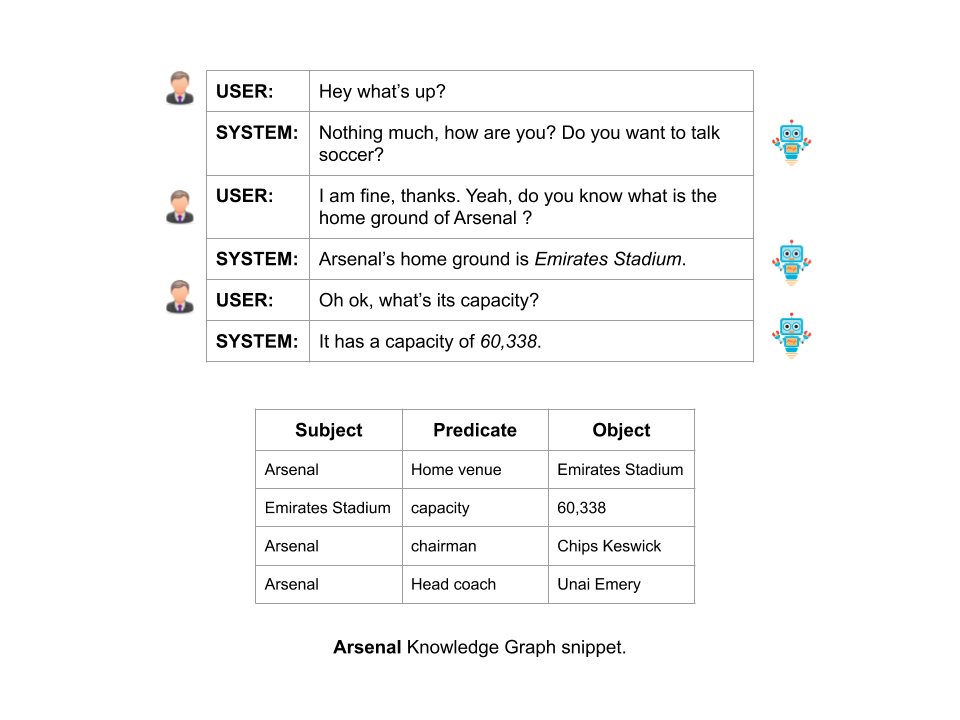}
        
    \caption{A conversation about the football club Arsenal and the Knowledge Graph involved.}
    \label{fig:convo}
    \vspace{0.1cm}
\end{figure}

For the case of non-goal oriented dialogues, the systems should be able to handle factoid as well as non-factoid queries like chit-chats or opinions on different subjects/domains. Generally, such systems are realized by using an extrinsic dialogue managers using intent detection subsequently followed by response generation (for the predicted intent) \cite{agrawal2018a}, \cite{akasaki-kaji-2017-chat}. Furthermore, in case of factoid queries posed to such systems, it is very important that they generate well articulated responses which are knowledge grounded. The systems must be able to generate a grammatically correct as well as factually grounded responses to such queries, while preserving co-references across the dialogue contexts. For better understanding, let us consider an example dialogue and the involved knowledge graph snippet in Figure~\ref{fig:convo}. The conversation consists of chit-chat as well as factoid queries. For the factoid question "do you know what is the home ground of Arsenal?", the system must be able to answer with the correct entity (Emirates Stadium) along with a grammatically correct sentence; as well as handle co-references("its" in the third user utterance meaning the stadium). Ideally, for an end-to-end system for non-goal oriented dialogues, the system should be able  to handle all these kind of queries using a single, end-to-end architecture. 

There are existing conversation datasets supported by knowledge graphs for well-grounded response generation. \cite{eric2017keyval}~introduced an in-car dialogue dataset for multi-domain, task-oriented dialogues along with a knowledge graph which can be used to answer questions about the task the user wants to be assisted with. The dataset consists of dialogues from the following domains: calendar scheduling, weather information retrieval, and point-of interest navigation.  For non-goal oriented dialogues, \cite{dodge2015evaluating}~proposed a dataset in the movie domain. The proposed dataset contains short dialogues for factoid question answering over movies or for recommendations. They also provide a knowledge graph consisting of triples as (s, r, o). Where s is the subject, r stands for relations and o being the object. An example of a triple from the dataset is: (Flags of Our Fathers, directed\_by, Clint Eastwood). The movie dialogues can utilize this provided knowledge graph for recommendation and question answering purposes. However, this dataset only tackles the problem of factual response generation in dialogues, and not well articulated ones.

To cater to the problem of generating well articulated, knowledge grounded responses for non-goal oriented dialogue systems, we propose a new dataset in the domain of soccer. We also propose the KG-Copy network which is able to copy facts from the KGs in case of factoid questions while generating well-articulated sentences as well as implicitly handling chit-chats, opinions by generating responses like a traditional sequence-to-sequence model.

The contributions of the paper can be summarized as follows:
\begin{itemize}
    \item A new dataset of 2,990 conversations for non-goal oriented dialogues in the domain of soccer, over various club and national teams.
    \item A soccer knowledge graph which consists of facts, as triples, curated from wikipedia.
    \item An end-to-end based, novel neural network architecture 
    as a baseline approach on this dataset. The network is empirically evaluated against other state-of-the-art architectures for knowledge grounded dialogue systems. The evaluation is done based on both knowledge groundedness using entity-F1 score and also standard, automated metrics (BLEU) for evaluating dialogue systems. 
\end{itemize}

The rest of the paper is organized as follows: we first introduce related work in Section 2. Then we cover the soccer dataset, which serves as background knowledge for our model in Section 3. The proposed model is explained in Section 4 and the training procedure is detailed in Section 5. In Section 6, we compare our model with other state-of-the-art models. We do a qualitative analysis of the model in Section 7, followed by an error analysis. In Section 8, finally we conclude.

\section{Related Work}
 Systems that are able to converse with humans have been one of the main focus of research from the early days of artificial intelligence.
 Such conversational systems can be designed as generative or retrieval based. A system produces automatic responses from the training vocabulary for the former, while selecting a best response from a set of possible responses for the latter.
 Automatic response generation was previously devised by \cite{ritter2011data} using a phrased-based generative method. 
Later onwards, sequence-to-sequence based neural network models has been mainly used for dialogue generation \cite{luong-etal-2015-addressing}, \cite{shang-etal-2015-neural}, \cite{Vinyals2015ANC}. 
These models are further improved using hierarchical RNN based architectures for incorporating more contextual information in the response generation process \cite{serban2016building}. Reinforcement learning-based end-to-end generative system were also proposed by~\cite{zhao2016towards} for jointly learning dialogue state-tracking \cite{williams2013dialog} and policy learning~\cite{baird1995residual}. 

\cite{lowe2015ubuntu} introduced the first multi-turn, retrieval based dataset which motivated a lot of further research on such systems. A lot of models are proposed on this dataset using both CNN \cite{yan2016learning}, \cite{an2018improving} and RNN \cite{wu2017sequential}, \cite{wang2015learning} based architectures.
Both generative and retrieval based models can benefit from additional world knowledge as mentioned previously. However, the task of incorporating such additional knowledge (both structured and unstructured) into dialogue systems is challenging and is also a widely researched topic. \cite{lowe2015incorporating}, \cite{xu2016incorporating}, \cite{chaudhuri2018improving} proposed architectures for incorporating unstructured knowledge into retrieval based systems. More recently, \cite{ghazvininejad2018knowledge} incorporated unstructured knowledge as facts into generative dialogue systems as well. 

Integration of structured knowledge comes in the form of incorporating knowledge graphs into the response generation process. \cite{eric2017keyval} proposed a Key-Value retrieval network along with the in-car dataset (consisting of goal-oriented dialogues) for KG integration into sequence-to-sequence model. \cite{madotto-etal-2018-mem2seq} proposed a generative model namely Mem2Seq for a task-oriented dialog system which combines multi-hop attention over memories with pointer networks. The model learns to generate dynamic queries to control the memory access. Mem2Seq is the current state-of-the-art on the in-car dataset. Further improvements on the task are proposed by \cite{kassawat2019jointemb} using joint embeddings and entity loss based regularization techniques. However, they learn the KG embeddings globally instead of per dialogue, so we evaluate our proposed system (KG-Copy network) against Mem2Seq. 

Alongside the previously mentioned datasets for knowledge grounded  dialogues, there is also a challenging dataset for complex sequential question answering which was introduced by~\cite{saha2018complex}. It contains around 200K sequential queries that require a large KG to answer. The dataset contains questions that require inference and logical reasoning over the KG to answer. Although the dataset is the first non-goal oriented dataset which aims at knolwedge graph integration, but it lacks proper conversational turns between utterances. 

\section{Soccer Dialogues Dataset} 

\subsection{Wizard-of-Oz Style Data Collection}

\begin{figure}
    \centering
    \includegraphics[width=\textwidth]{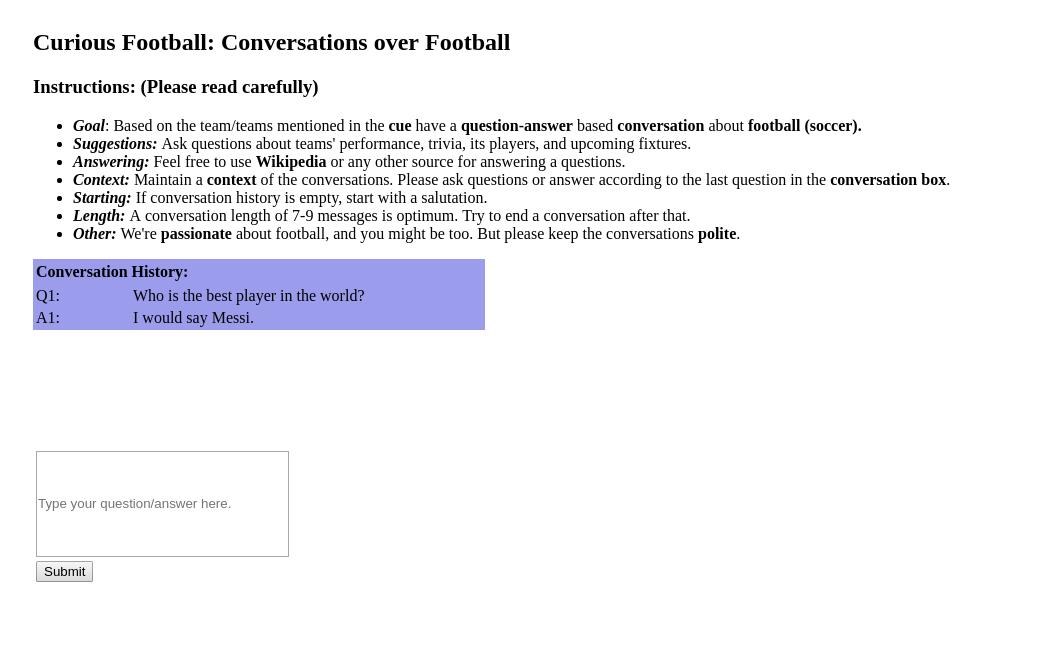}
       \vspace{-1cm}
    \caption{AMT setup for getting conversations over soccer.}
    \label{fig:amt}
    \vspace{-0.5cm}
\end{figure}

The proposed dataset for conversations over soccer is collected using AMT (Amazon Mechanical Turk)~\cite{doi:10.1177/1745691610393980}. The dialogues are collected in an wizard-of-oz style \cite{rieser2008learning} setup. In such a setup, humans believe they are interacting with machines, while the interaction is completely done by humans. 
The turkers, acting as users, were instructed to initiate a conversation about the given team with any query or opinion or just have some small-talks. This initial utterance is again posted as another AMT task for replying, this time a different turker is acting as a system. Turkers assigned to the system role were asked to use Wikipedia to answer questions posed by the user.
We encouraged the turkers to ask factual questions as well as posing opinions over the given teams, or have chit chat conversations. After a sequence of 7-8 utterances, the turkers were instructed to eventually end the conversation. A screenshot from the experimental setup is shown in Figure~\ref{fig:amt}.
We restricted the knowledge graph to a limited set of teams. 
The teams are picked based on popularity, the national teams chosen are: Sweden, Spain, Senegal, Portugal, Nigeria, Mexico, Italy, Iceland, Germany, France, Croatia, Colombia, Brazil, Belgium, Argentina, Uruguay and Switzerland. The club teams provided for conversing are: F.C. Barcelona, Real Madrid, Juventus F.C., Manchester United, Paris Saint Germain F.C., Liverpool F.C., Chelsea F.C., Atletico Madrid, F.C. Bayern Munich, F.C. Porto and Borussia Dortmund.  We also encouraged people to converse about soccer without any particular team. The number of conversations are equally distributed across all teams. The statistics of the total number of conversations are given in Table \ref{tab:conv}.

\begin{table*}[ht]
    \centering 
     \begin{tabular}{ l | c | c}
     \toprule
     \textbf{Dataset} & \textbf{ \# of Dialogues  } & \textbf{\# of Utterances} \\
     \hline
     Train & 2,493 & 12,243 \\
     Validation & 149 & 737 \\
     Test & 348 & 1,727 \\
     \bottomrule
    
    \end{tabular}
       \vspace{0.1cm}
     
    \caption{Statistics of Soccer Dataset.}
    \label{tab:conv}
    
\end{table*}

\subsection{Ensuring Coherence}

In order to ensure coherent dialogues between turkers, an additional task is created for each dialogue, where turkers were asked to annotate if the give dialogue is coherent or incoherent. Dialogues which are tagged incoherent by turkers are discarded. 

\subsection{Soccer Knowledge Graph}

A KG in the context of this paper is a directed, multi-relational graph that represents entities as nodes, and their relations as edges, which can be used as an abstraction of the real world. KGs consists of triples of the form (s,r,o) \begin{math}\in KG\end{math}, where s and o denote the subject and object entities, respectively, and r denotes their relation.

\begin{figure}
    \centering
    \includegraphics[width=\textwidth]{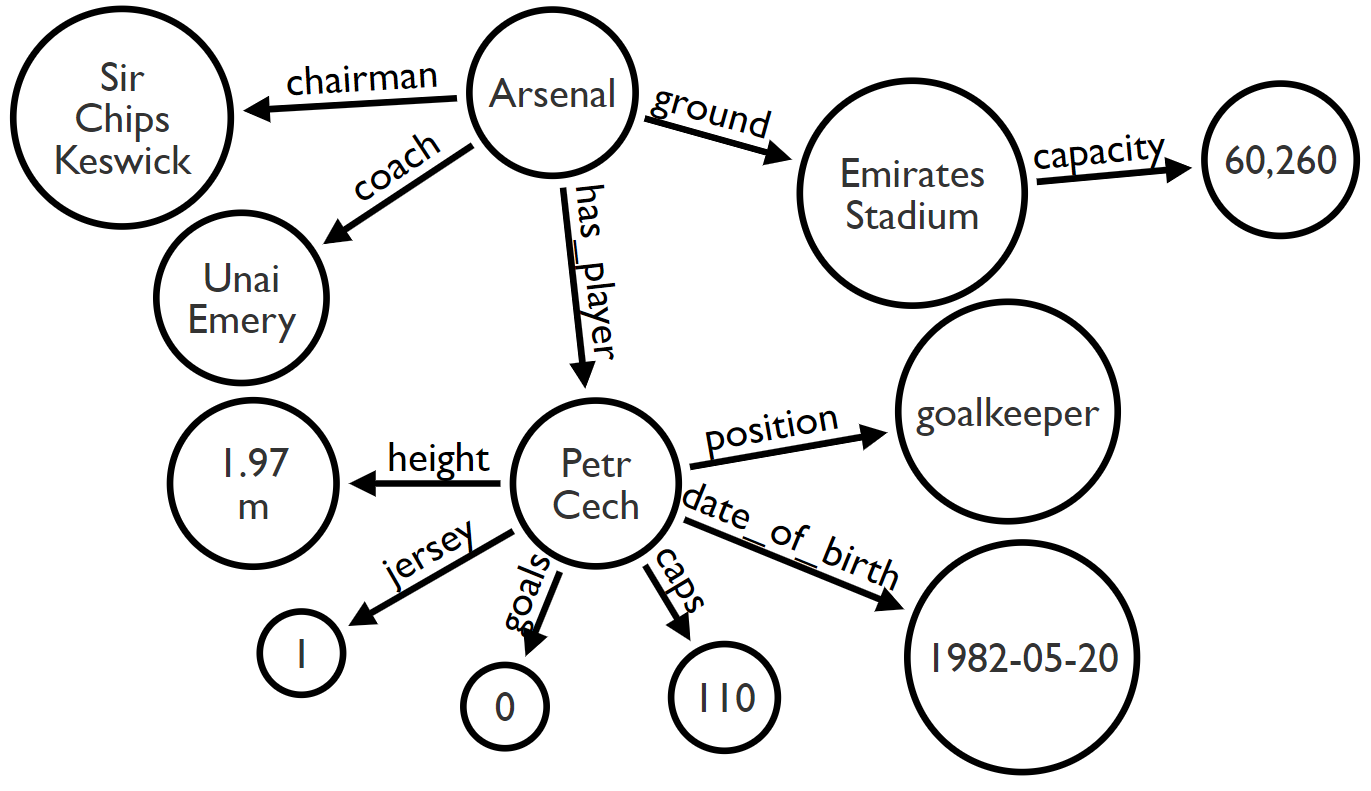}
        \vspace{0.1cm}
    \caption{Schema of the proposed Knowledge Graph for Arsenal.}
    \label{fig:schema}
    \vspace{0.1cm}
\end{figure}

Following~\cite{BergmannBEHKSS13a}, we created a soccer knowledge graph from WikiData \cite{Vrandecic:2014:WFC:2661061.2629489} which consists of information such as a team's coach, captain and also information such as home ground and its capacity for soccer clubs. For information about players, we have parsed individual wikipedia pages of the teams and mined goals scored, position, caps, height and age of players. This ensures that the info in the KG is up to date.
Finally, we curated the knowledge graphs for each team manually and added information such as jersey color. The  KG schema is provided in \ref{fig:schema} and additional statistics about KG and conversation is provided in table \ref{tab:kg_stat}. 

\begin{table*}[ht]
\vspace{0.1cm}
    \centering

     \begin{tabular}{ l | c }
     \toprule
     \textbf{Statistics} & \textbf{Count} \\
     \midrule
    Total Vocabulary Words ($v$) & 4782 \\
    Avg. Number of Conversations/team  & 83 \\
    Avg. Number of Triples/team & 148 \\
    Avg. Number of Entities/ team & 108 \\
    Avg. Number of Relations/team & 13 \\

     \bottomrule
    
    \end{tabular}
       \vspace{0.2cm}
    
    \caption{KG statistics.}
    \label{tab:kg_stat}
    
\end{table*}

 
\section{KG-Copy Model} 
The problem we are tackling in this paper is: given a knowledge graph (KG), and an input context in a dialogue, the model should be able to generate factual as well as well articulated response. During the dialogue generation process, at every time-step $t$, the model could either use the KG or generate a word from the vocabulary. We propose the KG Copy model which tackles this particular problem of producing well-grounded response generation.

KG-Copy is essentially a sequence-to-sequence encoder-decoder based neural network model, where the decoder can generate words either from the vocabulary or from the knowledge graph.
The model is mainly influenced by the copynets approach~\cite{gu2016copynet}. However, unlike copynets, KG-Copy copies tokens from the local knowledge graph using a special gating mechanism. Here, local KG depicts the KG for the team the dialogue is about. We introduce the KG-Copy's encoder, decoder and the gating mechanism below.

\subsection{KG-Copy Encoder}

The encoder is based on a recurrent neural network (RNN), more specifically a long-short term memory network (LSTM). It encodes the given input word sequence $X = [x_1, x_2..., x_T]$ to a fixed length vector $c$. The hidden states are defined by

\begin{equation}
    h_t = f_{enc}(x_t, h_{t-1})
\end{equation}

\noindent where $f_{enc}$ is the encoder recurrent function and the context vector $c$ is given by

\begin{equation}
    c = \phi(h_1, h_2...h_T)
\end{equation}

\noindent Here in, $\phi$ is the summarization function given the hidden states $h_t$. It can be computed by taking the last hidden state $h_T$ or applying attention over the hidden states~\cite{bahdanau2014neural,luong2015effective} and getting a weighted value of the hidden states (attention).

\subsection{KG-Copy Decoder}

The decoder is an attention based RNN (LSTM) model. The input to the decoder is the context $c$ from the encoder along with $h_T$. At time-step $t$, the hidden-state of the decoder is given by
\begin{equation}
    h^d_t = f_{dec}(x_{t}, h_{t-1})
\end{equation}

Where $f_{dec}$ is the recurrent function of the decoder. The decoder hidden-states are initialized using $h_T$ and the first token is <sos>. 
The attention mechanism \cite{luong2015effective}. The attention weights are calculated by concatenating the hidden states $h^d_t$ along with $h_t$ . 
\begin{equation}
    \alpha_t = softmax(W_s(tanh(W_c[h_t; h^d_t]))
\end{equation}
Here in, $W_c$ and $W_s$ are the weights of the attention model.
The final weighted context representation is given by  
\begin{equation}
    \Tilde{h_t} = \sum_t \alpha_t h_t
\end{equation}

 This representation is concatenated (represented by $;$) with the hidden states of the decoder to generate an output from the vocabulary with size $v$.
 
 The output is then given by 
 
 \begin{equation}
     o_t = W_o([{h_t; \Tilde{h^d_t}}])
 \end{equation}
 
 In the above equation, $W_o$ are the output weights with dimension $\mathbf{R}^{h_{dim}Xv}$. $h_{dim}$ is the dimension of the hidden layer of the decoder RNN. 
 
 \subsection{Sentient Gating}
 
 The sentient gating, as mentioned previously, is inspired mainly by \cite{gu2016copynet,merity2016pointer}. This gate acts as a sentinel mechanism which decides whether to copy from the local KG or to generate a word from training vocabulary ($v$). The final objective function can be written as the probability of predicting the next word during decoding based on the encoder hidden-states and the knowledge graph (KG).
 \begin{equation}
     p(y_t| h_t..h_1, KG)
 \end{equation}
 The proposed gating is an embedding based model. At every time-step $t$, the input query and the input to the decoder are fed into the sentient gate. Firstly, a simple averaging of the input query embedding is done generating $emb_q$, which can be treated as an vector representation of the input context.
\begin{equation}
    emb_q = \frac{1}{N}\sum(emb_{w1}....emb_{wt}) 
\end{equation} 

\noindent $emb_{wt}$ is the embedding of the $t^{th}$ word in the context. N.B. we only consider noun and verb phrases in the context to calculate $emb_q$. For the KG representation, an embedding average of the local KG's subject entity and relation labels for each triple is performed yielding a KG embedding $emb_{kg}$. We consider a total of $k$ triples in the local KG. 
 
 Finally, the query embedding is matched with these KG embeddings using a similarity function (cosine similarity in this case).
 \begin{equation}
     kg_{sim} = tanh(cos(emb_q, emb_{kg}^1), cos(emb_q, emb_{kg}^2)...cos(emb_q, emb_{kg}^k))
 \end{equation}
 
 The input to the decoder at $t$ is fed into the embedding too as mentioned previously yielding $emb_d$.
 
 The final sentient value at $t$ is given by :
 
 \begin{equation}
     s_t = sigmoid(W_{sent}[emb_q+emb_d; kg_{sim}; s_{t-1}])
 \end{equation} 
 
 $W_{sent}$ is another trainable parameter of the model and ";" is the concatenation operation. The final prediction is given by: 
 
 \begin{equation}
     out_t = s_t * kg_{sim} + (1 - s_t) * o_t
 \end{equation}

\begin{figure}
    \centering 
    \includegraphics[width=\textwidth]{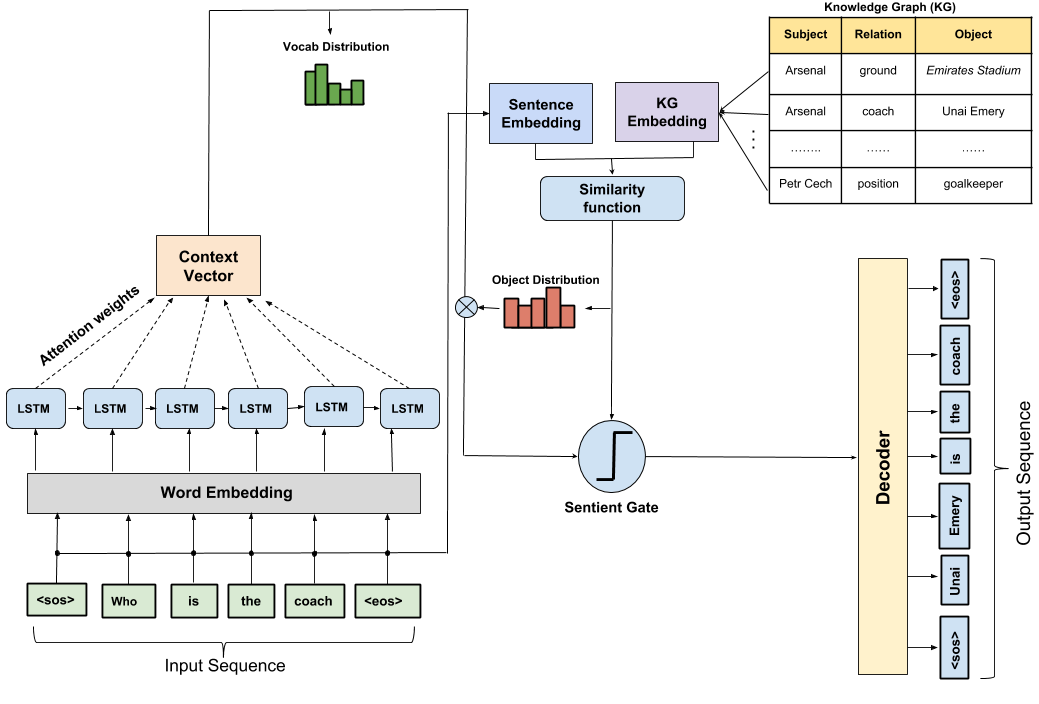}
        \vspace{-1cm}
    \caption{KG-Copy Model Encoder-Decoder Architecture for Knowledge Grounded Response Generation.}
    \label{fig:model}
    \vspace{0.1cm}
\end{figure}

The model is visualized in Figure \ref{fig:model}.

\section{Training and Model Hyper-parameters}

\subsection{Training Objective}

The model is trained based on a multi-task objective, where the final objective is to optimize the cross-entropy based vocabulary loss ($l_{vocab}$) and also the binary cross-entropy loss ($L_{sentient}$) for the sentient gate ($s_g$). This value is 1 if the generated token at that step comes from the KG, otherwise 0. For example, in the example provided in \ref{fig:convo}, for the $2^{nd}$ system utterance, this value would be 1 for $t=5$ (Emirates Stadium), but 0 for the previous time-steps.

\noindent The total loss is given by:

\begin{equation}
    L_{tot} = L_{vocab} + L_{sentient}
\end{equation}

\subsection{Training Details}

To train the model, we perform a string similarity over KG for each of the questions in training data set to find which questions are answerable from the KG. Then we replace those answers with the position number of the triples where the answer (object) belongs in the KG, during pre-processing. This is followed by a manual step where we verify whether the input query is simple, factoid question or not and also the correctness of answer (object). The vocabulary is built only using the training data. No additional pre-processing is done for the validation and test sets except changing words to their corresponding indices in the vocabulary.

For training, a batch-size of 32 is used and the model is trained for 100 epochs. We save the model with the best validation f1-score and evaluate {it on the test set.} We {apply Adam \cite{kingma2014adam} for optimization with a} learning rate of 1e-3 for the encoder and 5e-3 for the decoder. 
The {size of the hidden layer} of both the encoder and decoder LSTM is {set to} 64. We train the decoder RNN with teacher-forcing~\cite{williams1989learning}. The input word embedding layer is of dimension 300 and initialized with pretrained fasttext~\cite{bojanowski2017enriching} word embeddings. A dropout~\cite{srivastava2014dropout} of 0.3 is used for the encoder and decoder RNNs and 0.4 for the input embedding. The training process is {conducted} on a GPU with 3072 CUDA cores and a VRAM of 12GB. 
The soccer dataset (conversation and KG) and the KG-Copy model's code are open-sourced \footnote{https://github.com/SmartDataAnalytics/KG-Copy\_Network} for ensuring reproducibility.

\section{Evaluation} 

We compare our proposed model with Mem2Seq and a vanilla encoder-decoder with attention. We report the BLEU scores~\cite{papineni2002bleu} and also the entity-F1 scores on both the proposed soccer dataset and the In-car dialogue dataset. The results show that our proposed model performs better than both the vanilla attention sequence-to-sequence models and Mem2Seq model across both  metrics. Our model outperforms Mem2Seq by 1.51 in BLEU score and 15 \% on entity-F1 score. It performs better than the vanilla sequence-to-sequence model by 1.21 on the BLEU metric on the soccer dataset. Interestingly, Mem2Seq performs better than the vanilla model on validation, but it fails to generalize on test set. The proposed model although has lower BLEU on the in-car dialogue dataset, but has a better entity f1 scores (by 19.4 \%), implying stronger reasoning capabilities over entities and relations \cite{madotto-etal-2018-mem2seq}.  
\begin{table*}[ht]
    \centering 
    \vspace{-.4cm}
    \def\arraystretch{1.2}
     \begin{tabular}{ l | c | c}
     \toprule
     \textbf{Model} & \textbf{BLEU} & \textbf{Entity-F1} \\
     \hline
     
     & Valid  | Test  & Valid  | Test \\ \cline{2-2}\cline{3-3}
     Vanilla Encoder-decoder with Attention & 1.04 | 0.82 & \_ | \_ \\ 
     Mem2Seq \cite{madotto-etal-2018-mem2seq} & 1.30 | 0.52 & 6.78 | 7.03 \\
      \midrule
     KG Copy (proposed model) & \textbf{2.56} | \textbf{2.05} & 24.98 | 23.58 \\

     \bottomrule
    
    \end{tabular}
       \vspace{0.1cm}
     
    \caption{Results on Soccer Dataset.}
    \label{tab:test}
    
\end{table*}


    
     
    

\begin{table*}[ht]
\vspace{-0.1cm}
    \centering 
    \def\arraystretch{1.2}
     \begin{tabular}{ l | c | c}
     \toprule
     \textbf{Model} & \textbf{   BLEU  } & \textbf{Entity-F1}  \\
     \hline
     Vanilla Encoder-decoder with Attention & 8.4 & 10.3 \\
     Mem2Seq \cite{madotto-etal-2018-mem2seq} & \textbf{12.6} & 33.4 \\
      \midrule
     KG Copy (proposed model) & 9.6 & \textbf{52.8} \\

     \bottomrule
    
    \end{tabular}
       \vspace{0.1cm}
     
    \caption{Results on the In-car Dialogue Dataset.}
    \label{tab:test}
    
\end{table*}

\section{Discussion}

\subsection{Qualitative Analysis}
In this section, we will qualitatively analyze the response generation of our model along with the background knowledge integration (grounding) and compare it with both Mem2Seq and vanilla sequence-to-sequence models. 

Some example response from test are given in Table~\ref{tab:ground_resp}. As seen, the KG-copy model is able to have more articulate responses compare to sequence-to-sequence and Mem2Seq models. 
The model is also able to form well articulate opinions compared to other models ($2^{nd}$ column) \footnote{Seq2Seq model has generated a more articulated response based on the given context but it is factually wrong: Senegal is nicknamed the Lions of Teranga and not Nigeria.}.

\begin{figure}
    \centering 
    \includegraphics[width=\textwidth]{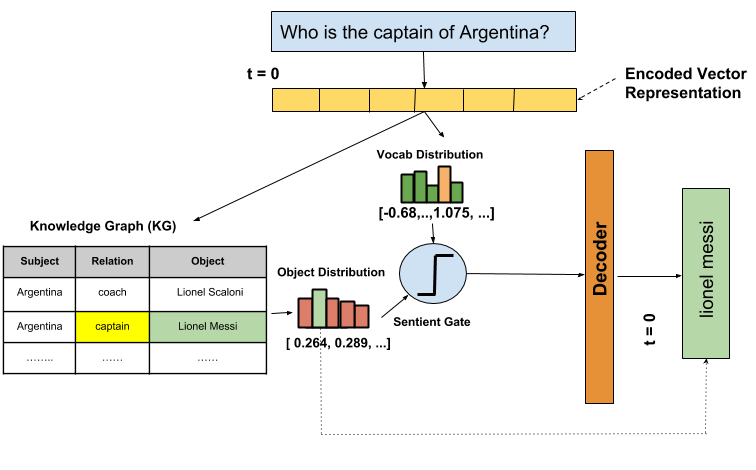}
        \vspace{-1cm}
    \caption{Response Generation during Decoding for KG-Copy Model.}
    \label{fig:decode}
    \vspace{-.1cm}
\end{figure}

Some more examples along with the response from our model are given in Table~\ref{tab:kg_copy_resp}. As observed, all those responses are well grounded. The first response is factually correct and also a well-articulate one; interestingly, even the true human response on the other hand is not. The last response is knowledge ground but not well articulate.
The model is also able to perform co-reference resolution implicitly while generating responses. To verify, let us consider another conversation between an user with the deployed KG-copy model. \\
\textbf{User utterance:} i like the team pretty much \\
\textbf{Response:} i don't think they're a lot of winning. \\
\textbf{User utterance:} who is the captain of argentina ? \\
\textbf{Response:} lionel messi is the captain \\
\textbf{User utterance:} do you know the name of their coach ? \\
\textbf{Response:} lionel scaloni is the coach\\
In the last response, the model is able to identify that the pronoun "their" refers to the team and is able to maintain a knowledge grounded, as well as articulate responses even for relatively long dialogue turns.
For time-step $t=0$, the visualization of the sentient gating mechanism is provided in Figure~\ref{fig:decode}. The vocabulary distribution is over $v$, and the object distribution here is over the local KG for the team. 

Furthermore, following~\cite{eric2017keyval}, we did an internal survey regarding the responses generated by KG-Copy network, judging the quality of responses based on the context on a scale of 1-5 on correctness and human-like sentence formation. The former measures how correct the generated response is with respect to the true response from the turker, and the latter how grammatically correct the produced response is. We randomly pick 50 conversation utterances from the test set and report this human evaluation both on Mem2Seq and KG-copy in Table~\ref{tab:human_eval}. 

\begin{table*}[ht]
    \centering 
   
\vspace{-.1cm}
     \begin{tabular}{ l | c | c  }
     \toprule
     \textbf{Model} & \textbf{Correctness} & \textbf{Human-like} \\
     \midrule
    Mem2Seq & 1.30 & 2.44 \\
    KG-Copy & 2.26 & 3.88 \\

     \bottomrule
    
    \end{tabular}
       \vspace{.1cm}
    
    \caption{Human evaluation of generated responses.}
    \label{tab:human_eval}
    
\end{table*}
\begin{table*}[ht]
    \centering 
   
\vspace{-.1cm}
     \begin{tabular*}{1.0\textwidth}{ l | p{46mm} |p{50mm}}
     \toprule
     \textbf{Context Type} & \textit{factoid} & \textit{opinions} \\
     \midrule
     \textbf{Input contexts} & what is the name of the captain of mexico ? & I like this team. \\
     \midrule
     \textbf{True response} & andres guardado (captain) & nigeria is a very well performing team and i like them a lot as well  \\
     \midrule
     \textbf{Seq2Seq} & The is is & They are nicknamed the Lions of Teranga.  \\
     \midrule
     \textbf{Mem2Seq  } & Mexico is the &  They are a\\
     \midrule
     \textbf{KG-Copy}  & andres guardado. & they are a good team.\\

     \bottomrule
    
    \end{tabular*}
       \vspace{.1cm}
    
    \caption{KG-copy Response for Factoid and non-Factoid queries.}
    \label{tab:ground_resp}
    
\end{table*}

\begin{table*}[ht]
    \centering 
   
\vspace{-.5cm}
     \begin{tabular*}{1.0\textwidth}{ p{40mm} | p{40mm} | p{40mm}}
     \toprule

     \textbf{input contexts} & \textbf{Turker Response} & \textbf{KG copy response} \\
     \midrule
    who is the captain of iceland? &aron gunnarsson &aron gunnarsson is the captain.\\ 
    \midrule
    who is the captain of italy ? & chiellini & giorgio chiellini is the captain.\\
    \midrule
    who is the coach for italy ? & i think roberto mancini & roberto mancini is the coach \\
    \midrule
    who is the coach of bayern munich ? & niko kovac is the current manager & niko kovac \\

     \bottomrule
    
    \end{tabular*}
       \vspace{0.1cm}
    
    \caption{KG Copy Model's Knowledge Grounded Responses.}
    \label{tab:kg_copy_resp}
    
\end{table*}

\subsection{Error Analysis}

Although the model is able to generate some well articulated, knowledge grounded responses for factual queries as seen in Tables~\ref{tab:ground_resp} and \ref{tab:kg_copy_resp},  the model often fails in producing factually correct responses as also evident from Table~\ref{tab:human_eval}. More of those cases are analyzed below. 

The model produces too generic and non-factual responses to queries about opinions about favorite players as shown in Table~\ref{tab:Wrong_opinion}. This is mostly because the vocabulary size is relatively large compare to the size of training dialogues. This can be improved with more training data, especially with more knowledge grounded conversations. For the first response, the model is not able to interpret the question and generates a generic response. For the second case, the generated response is not factual because the question is about Argentina, but Eden Hazard is from a different team (Belgium). 

\begin{table*}[ht]
    \centering 
   \vspace{-.1cm}

     \begin{tabular*}{1.0\textwidth}{ p{40mm} | p{30mm} | p{45mm}}
     \toprule
     \textbf{input contexts} & \textbf{True Response} & \textbf{KG copy response} \\
     \midrule
    who is senegal 's best current player not including mane ? & keita balde diao & i think it is the best player in the world cup \\
    \midrule
    who 's your favorite player ? & messi & i think eden hazard is the best player \\
     \bottomrule
    
    \end{tabular*}
       \vspace{.1cm}
    
    \caption{Incorrect opinionated responses from KG-Copy model.}
    \label{tab:Wrong_opinion}
    
\end{table*}

The KG-copy model also often suffers when more complex quantitative and other reasoning skills are required to respond to the context. For example, for the first context in Table~\ref{tab:complex}, the model needs to perform a count operation over the KG to answer it, which is currently unsupported. Similarly, for the second case the model would require better language inferencing to respond. The model also suffers from the problem of unknown words in the test set.

\begin{table*}[ht]
    \centering

     \begin{tabular*}{1.0\textwidth}{ l | p{50mm} | p{45mm}}
     \toprule
     \textbf{input contexts} & how many world cups has the brazil team won ? & who was the top scorer in the world cup for belgium ?\\
     \midrule
    \textbf{True Response}   & brazil has won the fifa world cup five times  & eden hazard \\
    \midrule
    \textbf{Predicted} & they won the world cup & i think it was the top scorer for the world cup \\
     \bottomrule
    
    \end{tabular*}
       \vspace{.1cm}
    
    \caption{Incorrect factual responses from KG Copy model.}
    \label{tab:complex}
    
\end{table*}

\section{Conclusion and Future Work}
In this paper, we introduce a new dataset for non-goal oriented, factual conversations over soccer (football). We also provide a knowledge graph for different club and national football teams which are the topic of these conversations.

Furthermore, we propose a relatively simple, novel, neural network architecture called KG-copy Network, as a baseline model, which can produce knowledge grounded responses as well as articulate responses via copying objects from the team KG based on the presented context of the question. Although the dataset is relatively small, the model can still learn the objective of producing grounded response as evident from the BLEU and entity-F1 scores compare to other models, and also from the examples provided in the paper. The proposed model also produces more knowledge grounded response (better entity f1 scores) on the in-car dialogue dataset \cite{eric2017keyval} compared to other approaches. However, it should be noted that the BLEU scores in case of the non-goal oriented soccer dataset is lower compare to the goal oriented dataset (in-car). This can be attributed to the fact that the vocabulary size in case of the former is much larger (3 times), hence proving it to be a much harder problem. We also outlined weaknesses and limitations, e.g.~for building factually correct responses, which can spur future research in this direction. 

As a future work, we would like to consider a bigger study for gathering more knowledge-grounded, non-goal oriented conversations extending to more domains other than soccer. One of the problem with the dataset is that some responses from the turkers themselves are not articulate enough as evident from Table~\ref{tab:kg_copy_resp}. To counter this, we would like to include more conversation verification steps and filter out conversations based on inter annotator agreements (IAA) between the turkers. Also, the proposed model can only respond to simple factoid questions based on word embedding based similarities between the context and the KG. We would like to extend the model to do better entity and relation linking between the query contexts and the knowledge graph in an end-to-end manner. The handling of out-of-vocabulary words also provides room for further research. Moreover, we would also like to investigate recently proposed transformer or BERT based sequence-to-sequence models for the task of knowledge grounded response generation.

\section{Acknowledgement}

This work has been supported by the Fraunhofer-Cluster of Excellence ``Cognitive Internet Technologies'' (CCIT).
%
%
\bibliographystyle{splncs04}
\bibliography{bibliography}
%

\end{document}